\pdfoutput=1

\documentclass[11pt]{article}

\usepackage[final]{acl}

\usepackage{times}
\usepackage{latexsym}

\usepackage[T1]{fontenc}

\usepackage[utf8]{inputenc}

\usepackage{microtype}

\usepackage{inconsolata}

\usepackage{graphicx}
\usepackage{booktabs}
\usepackage{tcolorbox}
\usepackage{listings}
\usepackage{multicol}
\usepackage{multirow}
\usepackage{array}
\usepackage{pifont}
\usepackage{amsmath}
\usepackage{enumitem}
\usepackage{colortbl}
\usepackage{subcaption}

\lstdefinestyle{plain}{
    basicstyle=\fontsize{7}{9.5}\ttfamily,
    keywordstyle=\color{blue},
    commentstyle=\color{gray},
    stringstyle=\color{green},
    showstringspaces=false,
    breaklines=true,
    breakatwhitespace=false,
    breakindent=0pt,
    escapeinside={(*@}{@*)}
}

\definecolor{MutedGreen}{RGB}{85, 170, 85}
\definecolor{CoolAccent}{RGB}{120, 145, 230}
\definecolor{IAP}{RGB}{255, 126, 121}
\definecolor{CoT}{RGB}{91, 155, 213}
\definecolor{WarmOrange}{RGB}{255, 165, 85}

\title{Steering the Compass: Aligning Dynamic Psychological Counseling Conversations with Cognitive Behavioral Therapy Strategies
}

\author{Zimu Wang$^{1,2}$, Yiwen Jiang$^{1}$, Xiangyu Zhao$^1$, Yaling Shen$^1$, Jiahe Liu$^1$, \\
  \textbf{Stephanie Fong$^{1}$, Maxmartwell H Cheng$^1$, Guilherme C Oliveira$^1$, } \\
  \textbf{Anh Nguyen$^2$, Robert Desimone$^3$, Barnaby Nelson$^{4,5}$, Dominic Dwyer$^{1,4,5}$, Zongyuan Ge$^{1}$} \\
  $^1$AIM for Health Lab, Monash University \ \ $^2$University of Liverpool \\
  $^3$Massachusetts Institute of Technology \ \ $^4$The University of Melbourne \ \ $^5$Orygen \\
  \texttt{Zimu.Wang@liverpool.ac.uk, Zongyuan.Ge@monash.edu}}

\begin{document}
\maketitle
\begin{abstract}
Recent advancements in large language models have revolutionized the field of psychological counseling, especially in the context of Cognitive Behavioral Therapy (CBT).
While the success of CBT relies heavily on \textit{dynamic decision-making} informed by the client's real-time mental state, this aspect has often been overlooked in current research, limiting both flexibility and therapeutic outcomes.
In this paper, we introduce \textsc{StratCBT}, a dataset specifically designed for psychological counseling conversations with \textbf{\underline{CBT}} \textbf{\underline{Strat}}egies, consisting of $9,688$ sessions and around $256$K utterances, with each counselor's response aligned with one of eight distinct strategies.
The creation of \textsc{StratCBT} involves modeling clients based on their negative thoughts and generating high-quality counseling conversations through self-chat, incorporating realistic sessions as guidance, thereby significantly surpassing existing datasets in both general counseling and CBT-specific skills.
We conduct extensive experiments to demonstrate the effectiveness of strategy-aligned generation and evaluate its efficacy in delivering professional and effective counseling with LLM-simulated clients to reflect real-world scenarios.
The dataset can be obtained from \url{https://github.com/zimuwangnlp/StratCBT}.
\end{abstract}

\section{Introduction}
\label{sec:intro}

Mental health is emerging as a global concern, affecting individuals across diverse ages and demographics \cite{na-etal-2025-survey,wang-etal-2025-posts}. Cognitive Behavioral Therapy (CBT) is a widely validated psychological treatment method that helps clients identify and correct negative and irrational thinking patterns through conversation \cite{na-2024-cbt,lee-etal-2024-cactus}. Recently, with the advent of large language models (LLMs), conversational CBT systems have become a key focus.
However, existing systems still face challenges with complex dialogue flows and personalized interventions, limiting their therapeutic efficacy and practical applicability.

\begin{figure}[t!]
    \centering
    \includegraphics[width=\linewidth]{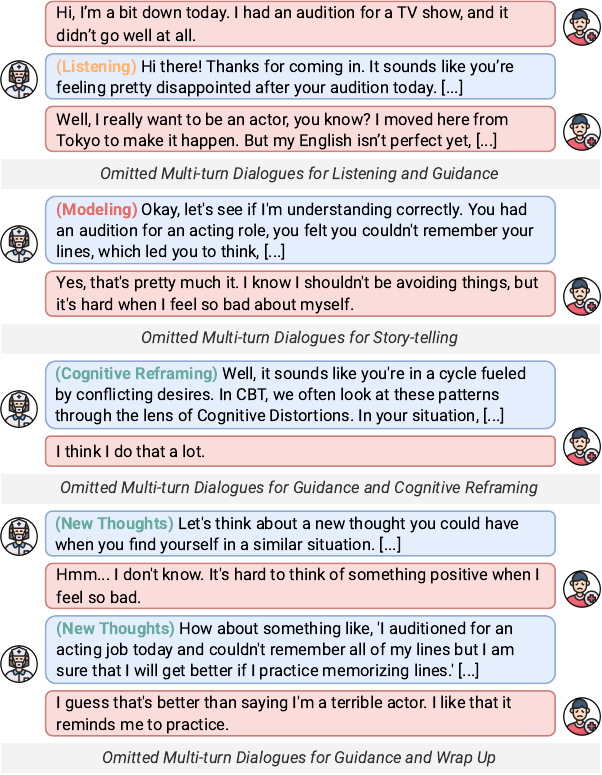}
    \caption{Example of a psychological counseling conversation with aligned CBT strategies.}
    \label{fig:example}
    \vspace{-4mm}
\end{figure}

The success of CBT interventions relies heavily on the \textit{dynamic decision-making} based on the client's state and reaction throughout the treatment process \cite{Turner1993,LEDDY2013173}. For instance, at certain points, the therapist may \textit{listen} and \textit{guide} the client to explore the events, thoughts, feelings, and behaviors that contribute to their negative emotions. At other times, the focus may shift towards \textit{cognitive restructuring} (see Figure \ref{fig:example}).
This dynamic adjustment ensures the coherence of the conversation and the gradual realization of therapeutic goals. Therefore, an effective CBT system must possess similar capabilities, assessing the client's real-time mental state and selecting the appropriate strategy to prevent the conversation from deviating from the treatment path or losing effectiveness.
However, while some studies have attempted to pre-plan the therapeutic process before the start of counseling dialogues, such as designing fixed conversation scripts or sequence strategies \cite{fu2023enhancing,lee-etal-2024-cactus}, such planning often lacks flexibility and is unable to respond to dynamic changes during the therapeutic process. Consequently, CBT systems need to move beyond static planning and adjust real-time strategies during the conversations.

Motivated by this gap, we introduce \textsc{StratCBT}, a dataset specifically designed for psychological counseling conversations with \textbf{\underline{CBT}} \textbf{\underline{Strat}}egies. It consists of $9,688$ sessions and around $256$K utterances, with each counselor's response aligned with one of eight distinct CBT strategies, such as \textit{Listening}, \textit{Modeling}, \textit{Cognitive Reframing}, and \textit{New Thoughts}.
Building on an initial set of $23$ collected CBT sessions, complete with counselor notes and the PatternReframe dataset \cite{maddela-etal-2023-training} tailored for cognitive restructuring, we engage an LLM to first model the clients based on their negative thoughts. This information is then utilized to generate high-quality counseling conversations through self-chat, incorporating realistic sessions to guide the generation process.
\textsc{StratCBT} surpasses existing datasets in both general counseling and CBT-specific skills.
We conduct experiments to address three core research questions:
the effectiveness of strategy-aligned generation,
the proficiency and preference of strategy prediction, and
the impact of strategy-aligned generation in multi-turn interactions with LLM-simulated clients.

The key contributions of this work are as follows:
\textbf{(1)} We emphasize strategy-aligned generation in CBT and introduce the \textsc{StratCBT} dataset, which possesses significantly better general counseling and CBT-specific skills;
\textbf{(2)} We propose an automatic framework for generating high-quality CBT sessions, leveraging existing cognitive restructuring datasets and self-chat;
\textbf{(3)} We conduct extensive experiments to demonstrate the effectiveness of strategy-aligned generation and explore its efficacy with LLM-simulated clients to reflect real-world scenarios.


\section{Related Work}

\begin{figure*}[t!]
    \centering
    \includegraphics[width=\linewidth]{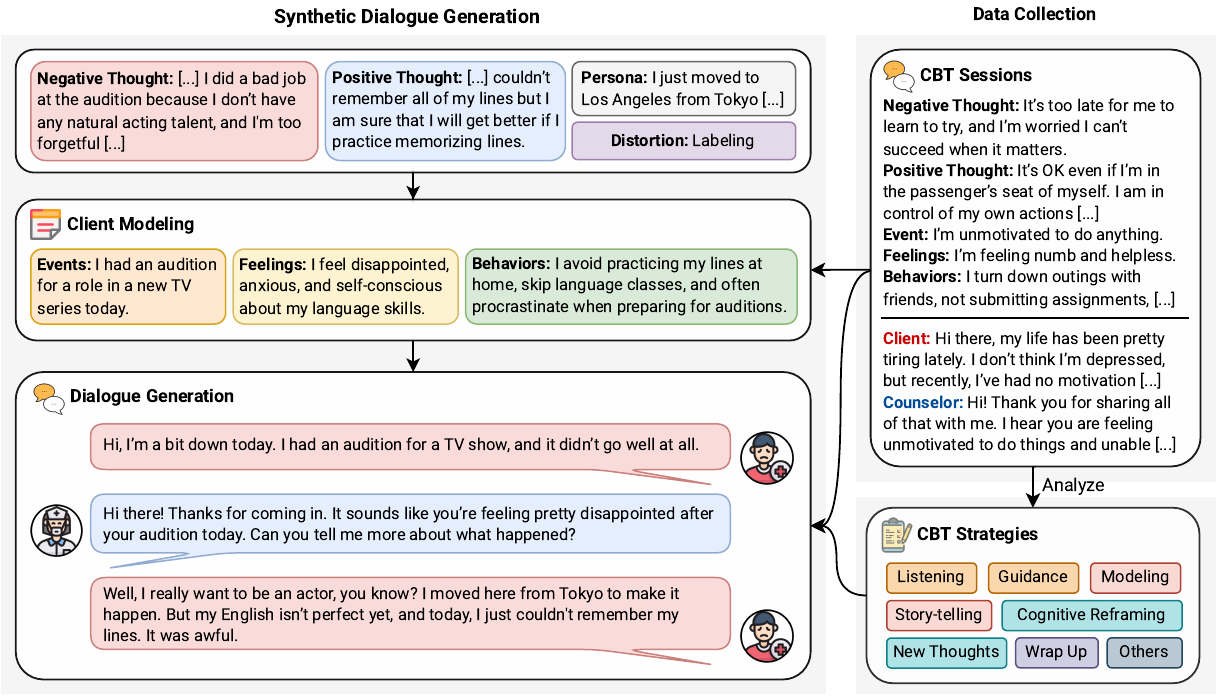}
    \caption{Overall construction pipeline for the \textsc{StratCBT} dataset.}
    \label{fig:data-collection}
    \vspace{-3mm}
\end{figure*}

\paragraph{LLM-based Psychological Counseling.}

The advancement of LLMs has broadened the potential for innovative mental health care due to their capability to generate human-like natural languages \cite{peng2023does,10580402}, offering promising applications across assessment \cite{wang-etal-2024-knowledge,ma-etal-2025-detecting,fong-etal-2026-chirpe}, diagnosis \cite{yao-etal-2022-d4,zhao2025hears}, and treatment \cite{na-etal-2026-overview,na-etal-2026-never,xu2026harm}.
In psychological counseling, ESConv \cite{liu-etal-2021-towards} and ExTES \cite{zheng-etal-2024-self} are introduced for emotional support conversations.
SoulChatCorpus \cite{chen-etal-2023-soulchat}, CPsyCounD \cite{zhang-etal-2024-cpsycoun}, and SMILE \cite{qiu-etal-2024-smile} are multi-turn counseling datasets derived from single-turn counseling cases.
Notably, \newcite{li-etal-2023-understanding} and \newcite{li-etal-2024-understanding-therapeutic} provide in-depth analysis of counselors' strategies, client reaction behaviors, and therapeutic relationships in online text-based counseling scenarios.
Several studies have also explored the integration of psychotherapy theories, such as CBT \cite{na-2024-cbt,xiao-etal-2024-healme,lee-etal-2024-cactus}, Motivational Interviewing (MI, \citealp{cohen-etal-2024-motivational,sun-etal-2024-eliciting}), and Solution-Focused Brief Therapy (SFBT, \citealp{chen2024mixed}).

However, existing works, particularly those for CBT, often rely on \textit{static} scripts between clients and therapists, which fail to align with therapeutic strategies and cannot capture clients' real-time mental state. While several emotional support studies have developed strategy schemes, these cannot be directly applied to CBT due to the distinct processes involved, such as \textit{modeling}, \textit{story-telling}, and \textit{cognitive reframing}. Different from previous work, we align the CBT counseling conversations with distinct strategies, enabling \textit{dynamic decision-making} throughout the therapeutic process.

\paragraph{Cognitive Behavioral Therapy.}

People with depression or anxiety often develop negative and irrational thinking patterns and reinforce negative beliefs \cite{Beck2011CognitiveBT}. CBT helps clients identify and correct these patterns through conversations.
With the emergence of LLMs, researchers have leveraged these models to perform key steps of CBT, such as cognitive restructuring \cite{maddela-etal-2023-training,sharma-etal-2023-cognitive,lin-etal-2024-detection}, and to facilitate both single-turn or multi-turn counseling conversations \cite{na-2024-cbt,xiao-etal-2024-healme,lee-etal-2024-cactus}.
In this work, we build on the CBT theory by introducing a comprehensive strategy scheme, the \textsc{StratCBT} dataset with the highest general counseling and CBT-specific skills, and the emphasis and analysis of dynamic strategy alignment in CBT counseling conversations.

\section{Dataset Construction}
\label{sec:data}

Figure \ref{fig:data-collection} presents the overall construction pipeline for the \textsc{StratCBT} dataset, including the collection of realistic CBT sessions ($\S$\ref{sec:collection}), the definition of the CBT strategy scheme ($\S$\ref{sec:cbt}), and the generation of synthetic dialogues ($\S$\ref{sec:synthetic}). In this section, we introduce each of the steps in detail.

\subsection{CBT Session Collection}
\label{sec:collection}

Existing psychological counseling datasets predominantly focus on dialogue generation using parametric knowledge from LLMs \cite{chen-etal-2023-soulchat,qiu-etal-2024-smile} or rely on multi-step prompting with handcrafted instructions for supervision \cite{xiao-etal-2024-healme,lee-etal-2024-cactus}. However, they often overlook the integration of real CBT sessions, leading to a gap with real-world therapeutic practice.
Motivated by this, we begin by collecting publicly available CBT sessions from CheeseBurger Therapy\footnote{\url{https://cheeseburgertherapy.org/}}, where each session is guided by counselors with $15-20$ hours of training in cognitive behavioral techniques, such as active listening, open-ended questioning, and cognitive restructuring, ultimately guiding clients in developing new, helpful thoughts.
Such trained peer counselors have been found to provide effective and empathetic CBT support \cite{10.1145/3613904.3642034,iftikhar2025therapy}, as evidenced by the client ratings following each session.
We collect $23$ sessions, each comprising the dialogue, the counselor's notes, and the new thoughts developed, covering a wide range of topics (e.g., self-growth, marriage, relationships) and participant backgrounds (e.g., students, workers).
These sessions are cleaned to maintain grammaticality and organized into one-to-one conversations.

\subsection{CBT Strategies}
\label{sec:cbt}

To provide coherent and effective CBT interventions, we develop a comprehensive strategy scheme based on CBT-related literature \cite{LEDDY2013173,iftikhar2025therapy}, consisting of $8$ distinct strategies: \textit{Listening}, \textit{Guidance}, \textit{Modeling}, \textit{Story-telling}, \textit{Cognitive Reframing}, \textit{New Thoughts}, \textit{Wrap Up}, and \textit{Others}.
The first two strategies create a warm and empathetic environment and foster clients' exploration of
their events, thoughts, feelings, and behaviors. The latter two strategies assist clients in categorizing and narrating their experiences, outlining how this cycle contributes to a negative feedback loop. The following two strategies guide clients in identifying cognitive distortions while creating new, positive thoughts. These strategies are employed interchangeably, facilitating a dynamic decision-making process. Detailed definitions of these strategies and their examples are shown in Appendix \ref{sec:strategies}. We manually annotate the counselors' utterances in the collected sessions, each based on its underlying therapeutic strategy.

\begin{table*}[t!]
    \centering
    \small
    \begin{tabular}{l|cccccc}
        \toprule
        \textbf{Dataset} & \textbf{Domain} & \textbf{Language} & \textbf{Strategy} & \textbf{\#Dialogue} & \textbf{\#Utterance} & \textbf{\#Avg. Utt.} \\
        \midrule
        ESConv \cite{liu-etal-2021-towards} & ES & English & \textcolor{ForestGreen}{\ding{51}} & $1,053$ & $31$K & $29.8$ \\
        ExTES \cite{zheng-etal-2024-self} & ES & English & \textcolor{ForestGreen}{\ding{51}} & $11,177$ & $200$K & $18.2$ \\
        \midrule
        SoulChatCorpus \cite{chen-etal-2023-soulchat} & GC & Chinese & \textcolor{BrickRed}{\ding{55}} & $2,300$K & $-$ & $-$ \\
        CPsyCoun \cite{zhang-etal-2024-cpsycoun} & GC & Chinese & \textcolor{BrickRed}{\ding{55}} & $3,134$ & $55$K & $17.6$ \\
        SmileChat \cite{qiu-etal-2024-smile} & GC & Chinese & \textcolor{BrickRed}{\ding{55}} & $55,165$ & $628$K & $11.4$ \\
        \midrule
        AnnoMI \cite{9746035} & MI & English & \textcolor{ForestGreen}{\ding{51}} & $110$ & $9$K & $80.3$ \\
        MI-TAGS \cite{cohen-etal-2024-motivational} & MI & English & \textcolor{ForestGreen}{\ding{51}} & $242$ & $16$K & $64.6$ \\
        BiMISC \cite{sun-etal-2024-eliciting} & MI & Bilingual & \textcolor{ForestGreen}{\ding{51}} & $80$ & $9$K & $109.4$ \\
        \midrule
        CBT-LLM \cite{na-2024-cbt} & CBT & Chinese & \textcolor{BrickRed}{\ding{55}} & $22,327$ & $45$K & $2.0$ \\
        HealMe \cite{xiao-etal-2024-healme} & CBT & English & \textcolor{BrickRed}{\ding{55}} & $1,300$ & $8$K & $6.0$ \\
        Cactus \cite{lee-etal-2024-cactus} & CBT & English & \textcolor{BrickRed}{\ding{55}} & $31,577$ & $964$K & $30.5$ \\
        \midrule
        \textsc{StratCBT} (Ours) & CBT & English & \textcolor{ForestGreen}{\ding{51}} & $9,688$ & $256$K & $26.4$ \\
        \bottomrule
    \end{tabular}
    \caption{Comparison between \textsc{StratCBT} and existing datasets. ``ES'' denotes emotional support, ``GC'' represents general counseling, and ``\#Avg. Utt.'' means the average number of utterances in each session.}
    \label{tab:statistics}
    \vspace{-1mm}
\end{table*}

\begin{table*}[t!]
    \centering
    \small
    \setlength{\tabcolsep}{2.1mm}{\begin{tabular}{c|l|ccc|ccc}
        \toprule
        \textbf{Evaluator} & \textbf{Dataset} & \textbf{Under.} & \textbf{Inter. Eff.} & \textbf{Colla.} & \textbf{Guided Dis.} & \textbf{Focus} & \textbf{Strategy} \\
        \midrule
        \multirow{7}{*}{GPT-4o} & ESConv \cite{liu-etal-2021-towards} & $3.11$ & $3.89$ & $2.98$ & $-$ & $-$ & $-$ \\
         & SoulChat \cite{chen-etal-2023-soulchat} & $4.55$ & $5.71$ & $3.96$ & $-$ & $-$ & $-$ \\
         & ExTES \cite{zheng-etal-2024-self} & $4.67$ & $5.40$ & $4.41$ & $-$ & $-$ & $-$ \\
         & CPsyCounD \cite{zhang-etal-2024-cpsycoun} & $4.89$ & $5.81$ & $4.73$ & $-$ & $-$ & $-$ \\
         & SMILE \cite{qiu-etal-2024-smile} & $4.97$ & $5.50$ & $4.76$ & $-$ & $-$ & $-$ \\
         & Cactus \cite{lee-etal-2024-cactus} & $\underline{5.76}$ & $\underline{5.94}$ & $\underline{5.60}$ & $\underline{5.00}$ & $\underline{5.70}$ & $5.37$ \\
         & \textsc{StratCBT} (Ours) & $\mathbf{5.84}^*$ & $\mathbf{5.95}$ & $\mathbf{5.79}^{**}$ & $\mathbf{5.18}^{***}$ & $\mathbf{5.99}^{***}$ & $\mathbf{5.77}^{***}$ \\
        \midrule
        Human & \textsc{StratCBT} (Ours) & $5.84$ & $5.94$ & $5.72$ & $5.16$ & $5.96$ & $5.84$ \\
        \bottomrule
    \end{tabular}}
    \caption{Data quality of \textsc{StratCBT} compared with existing dataset under six criteria: \textit{Understanding} (\textbf{Under.}), \textit{Interpersonal Effectiveness} (\textbf{Inter. Eff.}), \textit{Collaboration} (\textbf{Colla.}), \textit{Guided Discovery} (\textbf{Guided Dis.}), \textit{Focus}, and \textit{Strategy}. Significant tests are conducted over five runs ($^*$: $p<0.05$, $^{**}$: $p<0.01$, $^{***}$: $p<0.001$).}
    \label{tab:quality}
    \vspace{-2.4mm}
\end{table*}

\subsection{Synthetic Dialogue Generation}
\label{sec:synthetic}

\paragraph{Client Modeling.} The creation of CBT sessions begins by modeling the big picture of the clients that they are experiencing, including their \textit{events}, \textit{thoughts}, \textit{feelings}, and \textit{behaviors}, ensuring the realism and diversity of client backgrounds.
Without this process, clients may fail to pinpoint the specific events that trigger their negative emotions (e.g., responding with ``\textit{I'm not sure what has happened}'') or may provide overly simplistic or inconsistent responses throughout the conversation.

To achieve this, we utilize the PatternReframe dataset \cite{maddela-etal-2023-training}, which is tailored for cognitive restructuring and contains $9,688$ negative statements conditioned on a given persona, each paired with its cognitive distortion, and $26,507$ rewritten, complementary thoughts. For each negative statement paired with three reframed thoughts, we first utilize GPT-4o \cite{openai2024gpt4ocard} to select the most appropriate reframed thought that aligns with the client's persona and the given negative statement. This model is then employed to generate the events, feelings, and behaviors with respect to the negative thought and the client's persona. To better reflect real-world cases, in-context learning (ICL, \citealp{NEURIPS2020_1457c0d6}) is also employed, for which we select two in-context examples from the collected sessions with the highest semantic similarity to the negative thought, determined by the BGE-Large \cite{10.1145/3626772.3657878} embedding model.

\begin{figure}[t!]
    \centering
    \includegraphics[width=\linewidth]{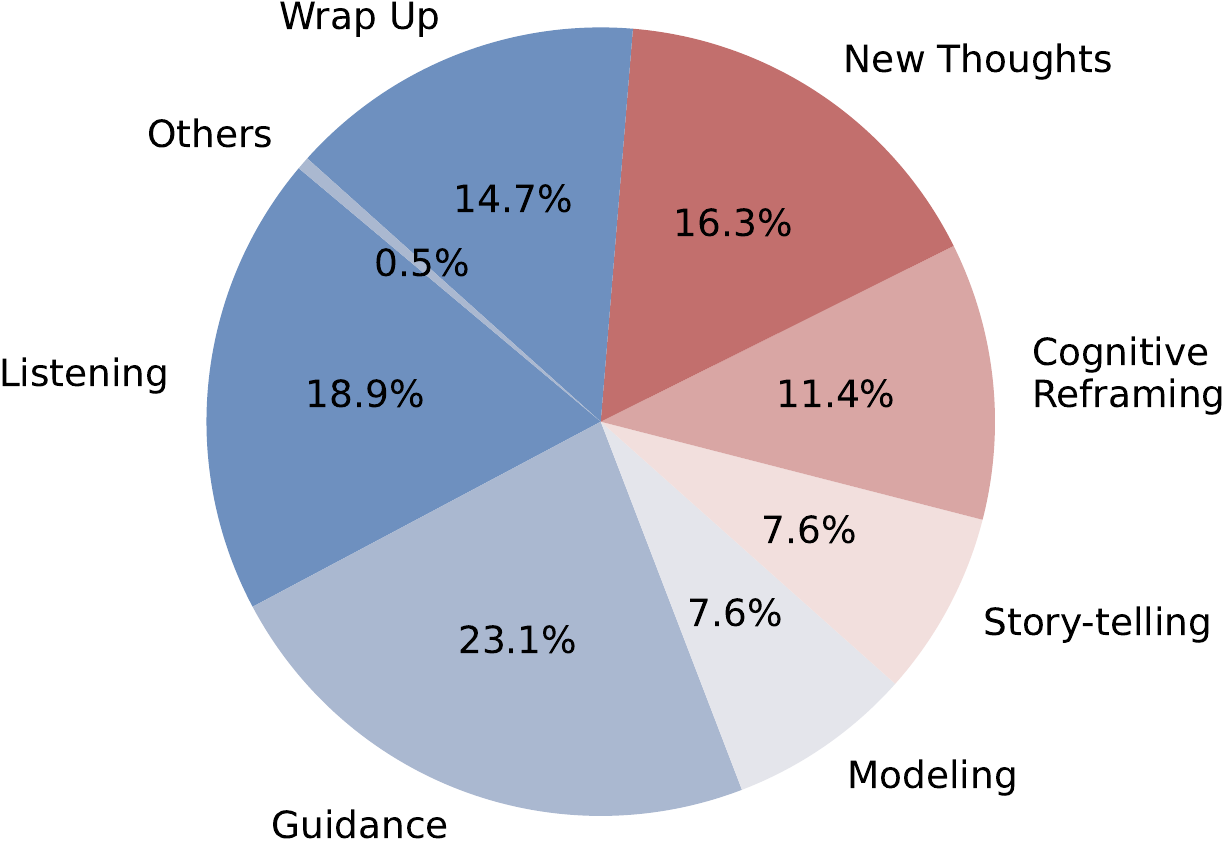}
    \caption{Strategy distribution in counselors' utterances.}
    \label{fig:strategy}
    \vspace{-4.2mm}
\end{figure}

\begin{figure*}[t!]
    \centering
    \includegraphics[width=\linewidth]{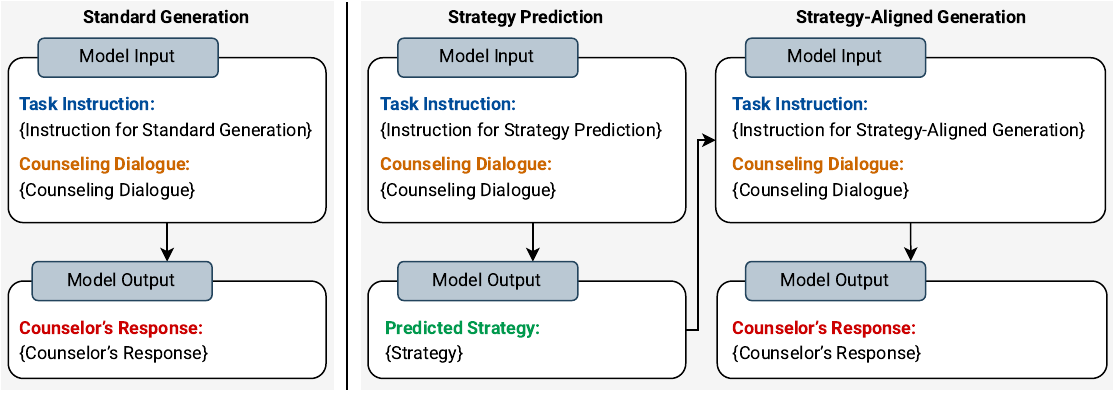}
    \caption{Illustration of the ``Standard Generation'' and the ``Strategy-Aligned Generation'' process.}
    \label{fig:setup}
    \vspace{-4mm}
\end{figure*}

\paragraph{Dialogue Generation.} In the final step, we generate counseling dialogues based on the modeled client information. Specifically, we prompt Gemini-2.0-Flash in a self-chat framework, which has been demonstrated to produce high-quality, multi-turn counseling conversations \cite{zhang-etal-2024-cpsycoun,zheng-etal-2024-self}.
Similar to the previous step, we leverage ICL with an in-context example selected based on the highest semantic similarity between the candidate client and the collected sessions in both the negative and restructured thoughts.
All relevant information, including the client's persona, cognitive distortion, and the events, thoughts, feelings, and behaviors, is provided, with detailed instructions outlining the strategies and how a CBT session should be structured (see Figure \ref{fig:generation-prompt}).
The model is then directed to generate the conversation, incorporating strategy annotations for the counselor's utterances to ensure alignment with CBT strategies.

\subsection{Dataset Characteristics}

\paragraph{Overall Statistics.}

Table \ref{tab:statistics} outlines the statistics of \textsc{StratCBT} in comparison to existing datasets, including emotional support, general counseling, MI, and CBT datasets, as the MI codes may also be considered a form of strategy.
\textsc{StratCBT} includes $9,688$ sessions and around $256$K utterances, averaging $26.4$ utterances in each session, which is consistent with previous datasets like Cactus \cite{lee-etal-2024-cactus}. However, it is the only dataset that aligns the counseling process with utterance-level CBT strategies.
As depicted in Figure \ref{fig:strategy}, the strategies of \textit{Guidance} and \textit{Listening} dominate the counselors' utterances, followed by \textit{New Thoughts}, \textit{Wrap Up}, and \textit{Cognitive Reframing}. This distribution closely mirrors real CBT scenarios, where counselors actively understand clients' backgrounds, engage in cognitive restructuring, and emphasize the intervention's effectiveness.

\paragraph{Data Quality.}

We perform a comprehensive comparison with existing datasets to highlight the quality of the \textsc{StratCBT} dataset. We select the Cognitive Therapy Rating Scale (CTRS), a gold standard for assessing the effectiveness of CBT conversations \cite{Aarons2012AdaptationHA}. Following \newcite{lee-etal-2024-cactus}, we select three criteria for evaluating general counseling skills (\textit{Understanding}, \textit{Interpersonal Effectiveness}, and \textit{Collaboration}) and three criteria for assessing CBT-specific skills (\textit{Guided Discovery}, \textit{Focus}, and \textit{Strategy}).
Evaluations are performed over five independent runs using GPT-4o, each with $100$ sessions, with prompts illustrated in Figures \ref{fig:ctrs-prompt} and \ref{fig:ctrs-prompt-2}. Human correlations of these metrics, as reported by \newcite{lee-etal-2024-cactus}, are presented in Table \ref{tab:correlation}.
To further confirm the validity of these metrics, we hired two postgraduate students in psychology to perform human evaluations on a subset of $100$ samples within our dataset.

As shown in Table \ref{tab:quality}, \textsc{StratCBT} demonstrates comparable performance in maintaining therapeutic relationships but significantly outperforms existing datasets across other criteria, including understanding the client's concerns, engaging collaborative goal-setting and decision-making, guiding client reflection and insight, addressing key cognitions and behaviors, and employing strategies to tackle problematic behaviors or thoughts. This underscores the effectiveness of the proposed strategy scheme and the \textsc{StratCBT} dataset in facilitating high-quality CBT counseling conversations.

\section{Experiments}

In this section, we present extensive experiments to address the following research questions (RQs):
\begin{itemize}[itemsep=0pt,topsep=0pt,parsep=0pt]
    \item \textbf{RQ1)} How effective is strategy-aligned generation in generating counseling responses?
    \item \textbf{RQ2)} What is the proficiency and preference of LLMs to predict the next CBT strategy used for response generation?
    \item \textbf{RQ3)} How does strategy-aligned generation impact multi-turn counseling conversations with LLM-simulated clients?
\end{itemize}

\subsection{Problem Definition}

We define the CBT counseling conversation task as follows: At each turn $t$, given the conversation history $\boldsymbol{u} = \{u_1^{cli}, u_1^{sys}, \ldots, u_t^{cli}\}$, the CBT system selects a strategy $s_t \in \mathcal{S}$, where $u^{cli}$ and $u^{sys}$ represent the client and system utterances, respectively, and $\mathcal{S}$ is a set of candidate CBT strategies (see $\S$\ref{sec:cbt}). Then, the CBT system generates the response $u_t^{sys}$ guided by the selected strategy $s_t$. For our experiments, we evaluate two generation approaches: \textit{Standard Generation} and \textit{Strategy-Aligned Generation}, as depicted in Figure \ref{fig:setup}. Prompts for standard generation, strategy prediction, and strategy-aligned generation are organized in Figure \ref{fig:conversation-prompt}.

\paragraph{Standard Generation.}

For standard generation, the next utterance $u_t^{sys}$ of the CBT system $\mathcal{M}_{SG}$ is generated solely based on the task definition $I_{SG}$ and the dialogue history $\boldsymbol{u}$, as outlined below:
\begin{equation}
    u_t^{sys} = \mathcal{M}_{SG}(\boldsymbol{u},I_{SG}).
\end{equation}

\paragraph{Strategy-Aligned Generation.}

Different from standard generation, strategy-aligned generation is dynamic to provide more specific information, ensuring that the generated response aligns with the selected strategy. Specifically, we first request a model $\mathcal{M}_{ST}$ to predict the next CBT strategy $s_t^{sys}$ from the set $\mathcal{S}$, which is then used to generate the system's utterance $u_t^{sys}$ with another model $\mathcal{M}_{SA}$:
\begin{align}
    s_t^{sys} &= \arg \max_{s \in \mathcal{S}} \mathcal{M}_{ST}(\boldsymbol{u}, I_{ST}), \label{eq:st} \\
    u_t^{sys} &= \mathcal{M}_{SA}(\boldsymbol{u}, D(s_t^{sys}), I_{SA}),
\end{align}
where $I_{ST}$ and $I_{SA}$ represent instructions for strategy prediction and strategy-aligned generation, and $D(s_t^{sys})$ denotes the definition of the selected strategy $s_t^{sys}$.

\begin{table*}[t!]
    \centering
    \small
    \begin{tabular}{l|cc|cc|cc|cc|cc|cc}
        \toprule
        \textbf{Model} & \multicolumn{2}{c|}{\textbf{BLEU}} & \multicolumn{2}{c|}{\textbf{ROUGE-L}} & \multicolumn{2}{c|}{\textbf{METEOR}} & \multicolumn{2}{c|}{\textbf{BERTScore}} & \multicolumn{2}{c|}{\textbf{Distinct-1}} & \multicolumn{2}{c}{\textbf{Distinct-2}} \\
        \cmidrule{2-13}
        \textit{Strategies} & w/o & w/ & w/o & w/ & w/o & w/ & w/o & w/ & w/o & w/ & w/o & w/ \\
        \midrule
        \rowcolor{gray!10}
        \multicolumn{13}{c}{\textit{Zero-shot LLMs}} \\
        \midrule
        \href{https://developers.openai.com/api/docs/models/gpt-4o}{GPT-4o} & $4.7$ & $\mathbf{6.4}$ & $22.2$ & $\mathbf{23.6}$ & $31.5$ & $\mathbf{33.6}$ & $87.4$ & $\mathbf{87.7}$ & $\mathbf{1.2}$ & $\mathbf{1.2}$ & $12.5$ & $\mathbf{12.9}$ \\
        \href{https://developers.openai.com/api/docs/models/gpt-4o-mini}{GPT-4o-mini} & $5.7$ & $\mathbf{7.3}$ & $23.2$ & $\mathbf{24.5}$ & $31.0$ & $\mathbf{33.8}$ & $87.8$ & $\mathbf{87.9}$ & $\mathbf{1.4}$ & $1.3$ & $\mathbf{13.0}$ & $12.5$ \\
        \href{https://ai.google.dev/gemini-api/docs/models/gemini-2.0-flash}{Gemini-2.0-Flash} & $9.1$ & $\mathbf{13.2}$ & $25.5$ & $\mathbf{28.7}$ & $32.9$ & $\mathbf{35.4}$ & $88.0$ & $\mathbf{88.6}$ & $1.5$ & $\mathbf{1.6}$ & $13.9$ & $\mathbf{14.0}$ \\
        \href{https://docs.mistral.ai/models/mistral-small-2-0-24-09}{Mistral-Small-2.0} & $6.2$ & $\mathbf{7.7}$ & $21.7$ & $\mathbf{23.4}$ & $29.3$ & $\mathbf{32.4}$ & $\mathbf{87.4}$ & $\mathbf{87.4}$ & $\mathbf{1.4}$ & $1.3$ & $\mathbf{12.7}$ & $12.1$ \\
        \href{https://ai.meta.com/blog/llama-4-multimodal-intelligence/}{Llama-4-Maverick} & $5.5$ & $\mathbf{8.1}$ & $22.1$ & $\mathbf{23.9}$ & $30.7$ & $\mathbf{33.6}$ & $87.3$ & $\mathbf{87.7}$ & $\mathbf{1.3}$ & $\mathbf{1.3}$ & $\mathbf{12.3}$ & $12.2$ \\
        \href{https://api-docs.deepseek.com/news/news250325/}{DeepSeek-V3} & $2.8$ & $\mathbf{4.6}$ & $17.2$ & $\mathbf{19.8}$ & $28.3$ & $\mathbf{30.8}$ & $85.5$ & $\mathbf{86.3}$ & $\mathbf{1.6}$ & $1.5$ & $13.6$ & $\mathbf{14.9}$ \\
        \midrule
        \rowcolor{gray!10}
        \multicolumn{13}{c}{\textit{Fine-tuned LLMs}} \\
        \midrule
        \href{https://ai.meta.com/blog/meta-llama-3-1/}{Llama3.1-8B} & $22.3$ & $\mathbf{23.6}$ & $36.3$ & $\mathbf{37.2}$ & $40.4$ & $\mathbf{40.9}$ & $90.4$ & $\mathbf{90.5}$ & $\mathbf{2.1}$ & $\mathbf{2.1}$ & $\mathbf{16.5}$ & $16.3$ \\
        \href{https://docs.mistral.ai/models/ministral-8b-24-1}{Ministral-8B} & $22.9$ & $\mathbf{23.5}$ & $36.8$ & $\mathbf{37.4}$ & $41.0$ & $\mathbf{41.5}$ & $90.5$ & $\mathbf{90.6}$ & $\mathbf{2.0}$ & $\mathbf{2.0}$ & $15.9$ & $\mathbf{16.1}$ \\
        \href{https://falconllm.tii.ae/falcon3/index.html}{Falcon3-7B} & $22.9$ & $\mathbf{23.5}$ & $36.8$ & $\mathbf{37.3}$ & $40.5$ & $\mathbf{40.9}$ & $\mathbf{90.5}$ & $\mathbf{90.5}$ & $\mathbf{2.0}$ & $\mathbf{2.0}$ & $15.9$ & $\mathbf{16.0}$ \\
        \href{https://qwen.ai/blog?id=qwen3}{Qwen3-4B} & $22.6$ & $\mathbf{23.4}$ & $36.0$ & $\mathbf{36.7}$ & $40.1$ & $\mathbf{40.7}$ & $90.4$ & $\mathbf{90.5}$ & $\mathbf{2.0}$ & $\mathbf{2.0}$ & $\mathbf{15.7}$ & $15.5$ \\
        \href{https://qwen.ai/blog?id=qwen3}{Qwen3-8B} & $23.0$ & $\mathbf{23.6}$ & $36.6$ & $\mathbf{37.2}$ & $40.5$ & $\mathbf{40.9}$ & $\mathbf{90.5}$ & $\mathbf{90.5}$ & $2.0$ & $\mathbf{2.1}$ & $16.1$ & $\mathbf{16.3}$ \\
        \bottomrule
    \end{tabular}
    \caption{Experimental results on \textsc{StratCBT} without (w/o) and with (w/) strategy-aligned generation. The better performance for each model and metric is highlighted in \textbf{bold}.}
    \label{tab:results}
    \vspace{-2mm}
\end{table*}

\begin{table}[t!]
    \centering
    \small
    \begin{tabular}{l|ccc}
        \toprule
        \textbf{Model} & \textbf{BLEU} & \textbf{ROUGE-L} & \textbf{BERTScore} \\
        \midrule
        \rowcolor{gray!10}
        \multicolumn{4}{c}{\textit{GPT-4o-mini}} \\
        \midrule
        Standard & $5.7$ & $\underline{23.2}$ & $\underline{87.8}$ \\
        w/ Direct-Refine & $4.1$ & $21.1$ & $87.4$ \\
        w/ Self-Refine & $3.5$ & $19.9$ & $86.7$ \\
        w/ Knowledge & $5.4$ & $22.7$ & $\underline{87.8}$ \\
        Strategy-Aligned & $\underline{7.3}$ & $\mathbf{24.5}$ & $\mathbf{87.9}$ \\
        \midrule
        \rowcolor{gray!10}
        \multicolumn{4}{c}{\textit{Gemini-2.0-Flash}} \\
        \midrule
        Standard & $9.1$ & $\underline{25.5}$ & $\underline{88.0}$ \\
        w/ Direct-Refine & $7.5$ & $24.0$ & $87.6$ \\
        w/ Self-Refine & $8.1$ & $24.7$ & $87.8$ \\
        w/ Knowledge & $\underline{9.2}$ & $25.1$ & $\underline{88.0}$ \\
        Strategy-Aligned & $\mathbf{13.2}$ & $\mathbf{28.7}$ & $\mathbf{88.6}$ \\
        \bottomrule
    \end{tabular}
    \caption{Results of strategy-aligned generation against baselines. The best and the second-best performance for each model and metric are in \textbf{bold} and \underline{underlined}.}
    \label{tab:baselines}
    \vspace{-4mm}
\end{table}

\subsection{Counseling Response Generation (RQ1)}

\paragraph{Models and Setup.}
We conduct experiments in both zero-shot and fine-tuning settings using state-of-the-art LLMs.
For zero-shot prompting, we experiment with GPT-4o (\texttt{2024-08-06}), GPT-4o-mini, Gemini-2.0-Flash, Mistral-Small-2.0, Llama-4-Maverick, and DeepSeek-V3 (\texttt{0324}), and we fine-tune models including Llama3.1-8B, Ministral-8B, Falcon3-7B, Qwen3-4B, and Qwen3-8B with LoRA \cite{hu2022lora}.
While these models are instruction-tuned, we conduct an analysis using GPT-4o-mini to explore their capability for internal mental state reasoning in comparison to explicit Chain-of-Thought (CoT, see Appendix \ref{sec:cot}).
During the fine-tuning process, we set the number of epochs to $10$, the batch size to $1$, the gradient accumulation step to $8$, and the learning rate to $1e-4$. All experiments are conducted on $4$ NVIDIA GeForce RTX 3090 graphics cards.

\paragraph{Baselines.}

To evaluate the efficacy of strategy-aligned generation, we conduct a comparative analysis with well-established baselines in the domains of dialogue systems and emotional support \cite{NEURIPS2023_91edff07,cai-etal-2024-empcrl,kang-etal-2024-large}, including (1) \textbf{Direct-Refine}, which involves the model self-defining and refining its initial response; (2) \textbf{Self-Refine}, which iteratively refines the response based on self-feedback; and (3) \textbf{Knowledge-Enhanced Generation}, which augments the generation process by incorporating commonsense causal knowledge relevant to the ongoing dialogue. Detailed descriptions of these baselines can be found in Appendix \ref{sec:baselines}.

\paragraph{Evaluation Metrics.}

We employ the following metrics to evaluate the generation quality objectively on a subset of $500$ dialogues from the test set: (1) \textbf{BLEU} \cite{papineni-etal-2002-bleu} and \textbf{ROUGE-L} \cite{lin-2004-rouge} assess the $n$-gram overlap between the generation and the reference; (2) \textbf{METEOR} \cite{banerjee-lavie-2005-meteor} evaluates the semantic and syntactic accuracy; (3) \textbf{BERTScore} \cite{Zhang2020BERTScore} calculates semantic similarity using BERT embeddings; (4) \textbf{Distinct-1} and \textbf{Distinct-2} assess the lexically diversity based on the unique $n$-grams ($n=1,2$) in the generated content.

\paragraph{Evaluation Results.}

Tables \ref{tab:results} and \ref{tab:baselines} illustrate the results of the evaluated models, which consistently demonstrate notable improvements when strategies are applied.
Additionally, the diversity of the models remains comparable or even improves, as reflected in the Distinct-1 and Distinct-2 scores.

Employing large-scale models through zero-shot prompting does not yield satisfactory results;
however, strategy-aligned generation benefits all models and outperforms all baselines, particularly direct-refine and self-refine.
By manually analyzing the feedback from self-refine, we find that these baselines primarily emphasize \textit{empathy} in responses (e.g., ``\textit{could enhance its empathic engagement}'').
While this focus is effective in emotional support conversations, it is less suitable for professional CBT counseling, where a more nuanced, task-specific approach is required.
Under fine-tuning, the models show similar performance upon aligning the response with the predicted strategies. Notably, the performance gap between Qwen3-4B and Qwen3-8B is diminished with strategy-aligned generation. This underscores its effectiveness in reducing performance discrepancies across different model architectures and parameter scales.

\subsection{Strategy Prediction and Preference (RQ2)}

\paragraph{Evaluation Metrics.} We evaluate both the \textit{proficiency} and \textit{preference} of LLMs in their ability to select the appropriate strategy and their preference towards certain strategies over others. The proficiency in strategy prediction ($q_{s_i}$) is quantified as the accuracy for each individual strategy $s_i$,  while an overall proficiency, $\mathcal{Q}$, is computed to reflect the model's performance across all strategies.

In preference evaluation ($p_{s_i}$), we follow \newcite{kang-etal-2024-large} in employing the Bradley-Terry model \cite{19ff28b9-64f9-3656-ba40-08326a05748e}.
Formally, the preference for strategy $s_i$ can be derived as follows:
\begin{equation}
    p_{s_i}' = \frac{\sum_j (w_{ij} \cdot p_{s_j}) / (p_{s_i} + p_{s_j})}{\sum_j w_{ji} / (p_{s_i} + p_{s_j})},
    \label{eq:bt}
\end{equation}
where $w_{ij}$ denotes the frequency with which the model predicts strategy $s_i$ when the ground-truth strategy is $s_j$. All preferences $p_{s_i}$ are initialized as 1 and updated through the iterative application of Equation (\ref{eq:bt}), where $p'_{s_i}$ is the preference in the next iteration.
After the final iteration, the sum of all $p_i$ values is normalized to $8$, ensuring the average $p$ of $1$. A value of $p_{s_i} > 1$ thus indicates a strong preference for $s_i$. Furthermore, we calculate a standard deviation of preferences across strategies, which we define as the preference bias $\mathcal{B}$, as follows:
\begin{equation}
    \mathcal{B} = \sqrt{\frac{\sum_{i=1}^N (p_i - \bar{p})^2}{N}},
\end{equation}
where a higher value of $\mathcal{B}$ indicates a distinct preference for both preferred and non-preferred strategies. Details of the Bradley-Terry model and the evaluation are explained in Appendix \ref{sec:bt}.

\paragraph{Experimental Results.}

\begin{figure}[t!]
    \centering
    \includegraphics[width=\linewidth]{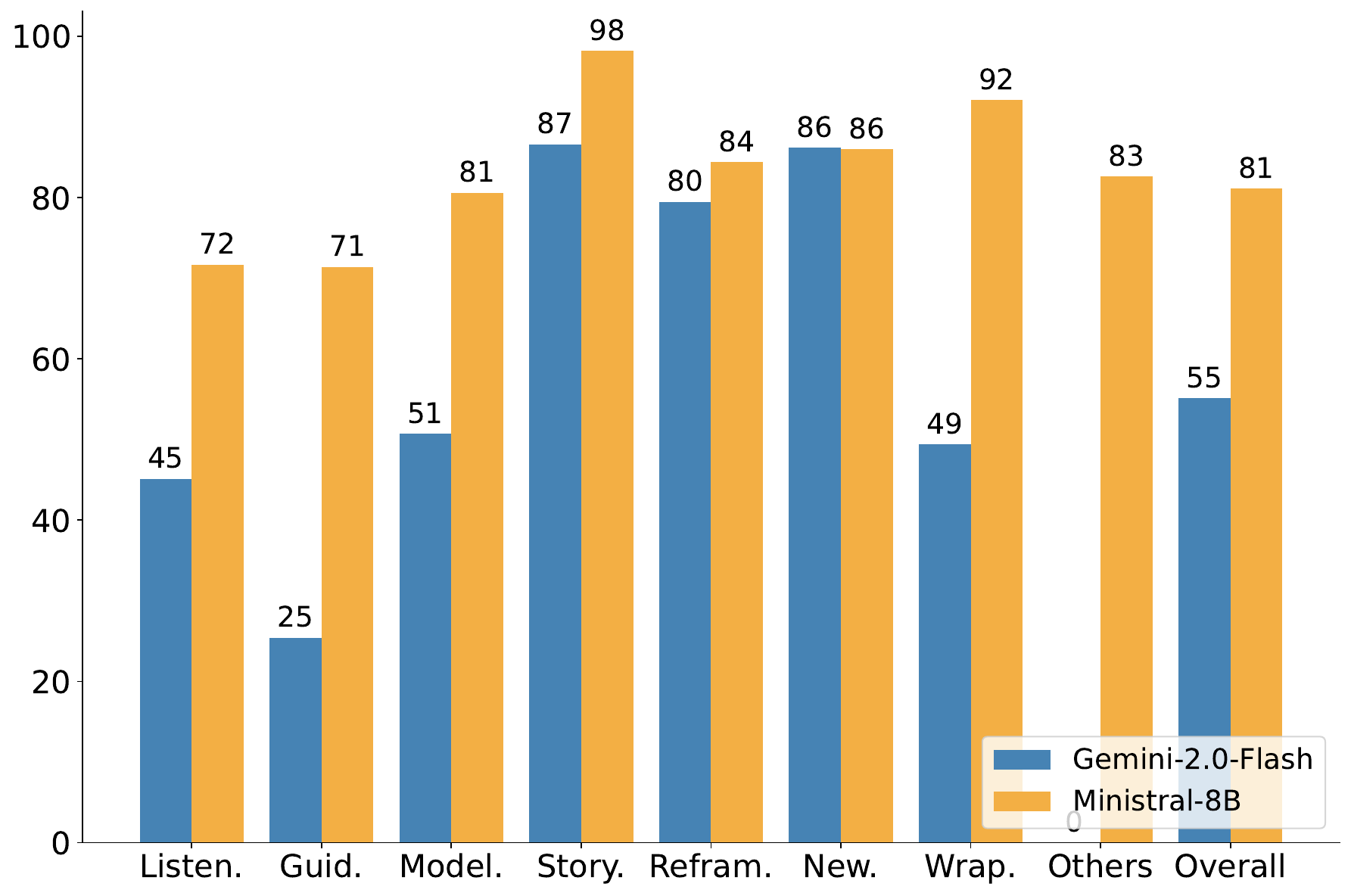}
    \includegraphics[width=\linewidth]{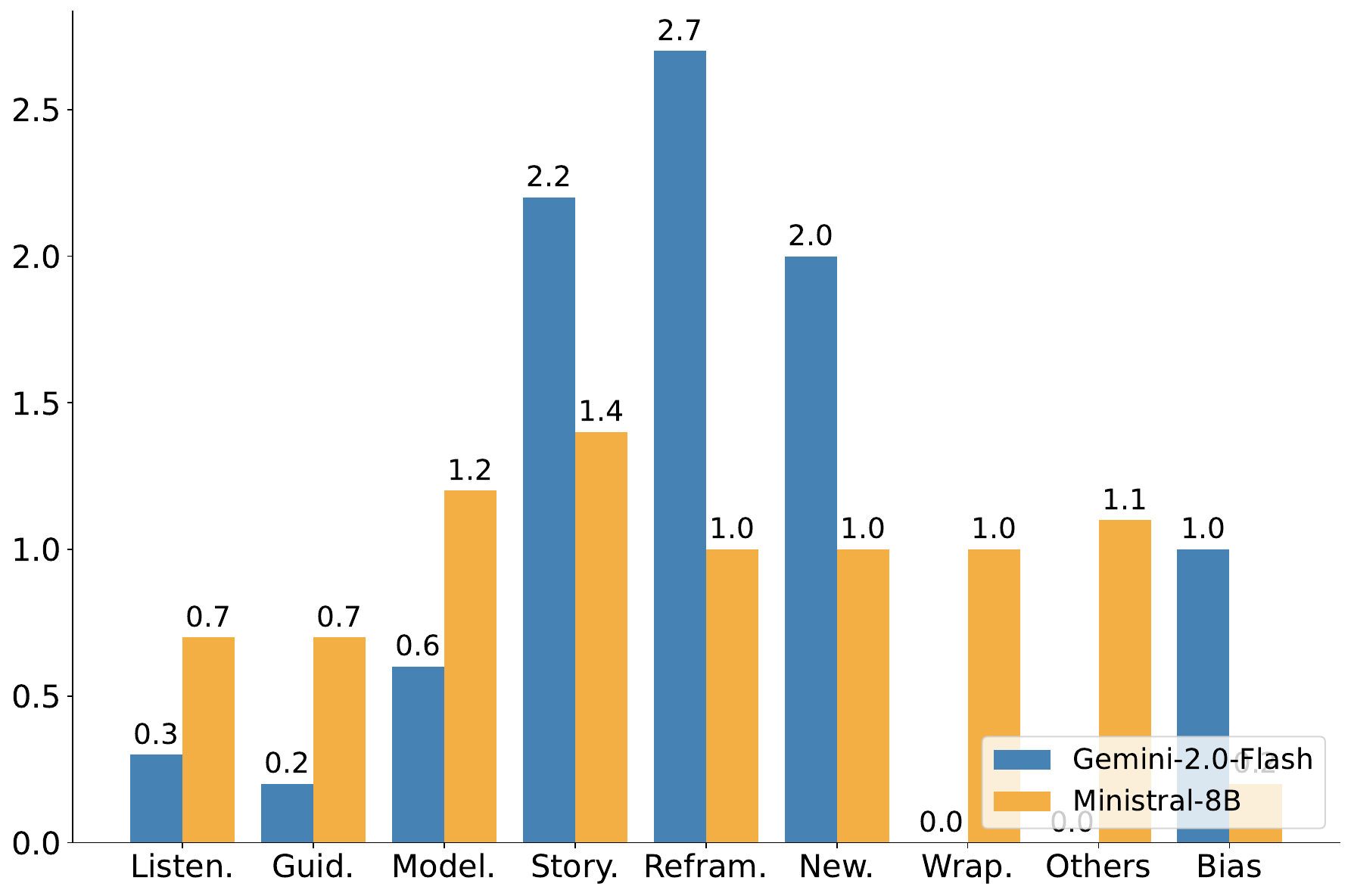}
    \caption{Proficiency (\textit{top figure}) and preference (\textit{bottom figure}) of Gemini-2.0-Flash and Ministral-8B on strategy prediction. Full results are in Table \ref{tab:pred_strategy}.}
    \label{fig:proficiency}
    \vspace{-4mm}
\end{figure}

Figure \ref{fig:proficiency} illustrates the proficiency and preference of the best-performed zero-shot and fine-tuned models.
The strong correlation between strategy prediction and response generation results underscores the significant role that strategies play in CBT conversations.
Zero-shot models typically prioritize strategies such as \textit{Cognitive Reframing} and \textit{New Thoughts}, recognizing their key role within CBT. However, these models often overlook the broader context that underpins the client's negative emotions. Following fine-tuning, while the models continue to perform well in predicting the core strategies, they witness a notable improvement in their ability to predict additional strategies, particularly \textit{Listening} and \textit{Guidance}, showcasing a robust generalization of the models across strategies.

\subsection{Simulated Counseling Scenarios (RQ3)}

To further validate strategy-aligned generation in real-world therapeutic contexts, we conduct client simulation using $20$ samples selected from the Patient-$\Psi$-DM \cite{wang-etal-2024-patient} dataset.
Each client sample consists of eight components, such as relevant history and core belief. We employ GPT-4o to simulate these clients and utilize GPT-4o and strategy-aligned generation based on Ministral-8B (Ministral-8B-SA) as counselors.

\begin{figure}[t!]
    \centering
    \includegraphics[width=\linewidth]{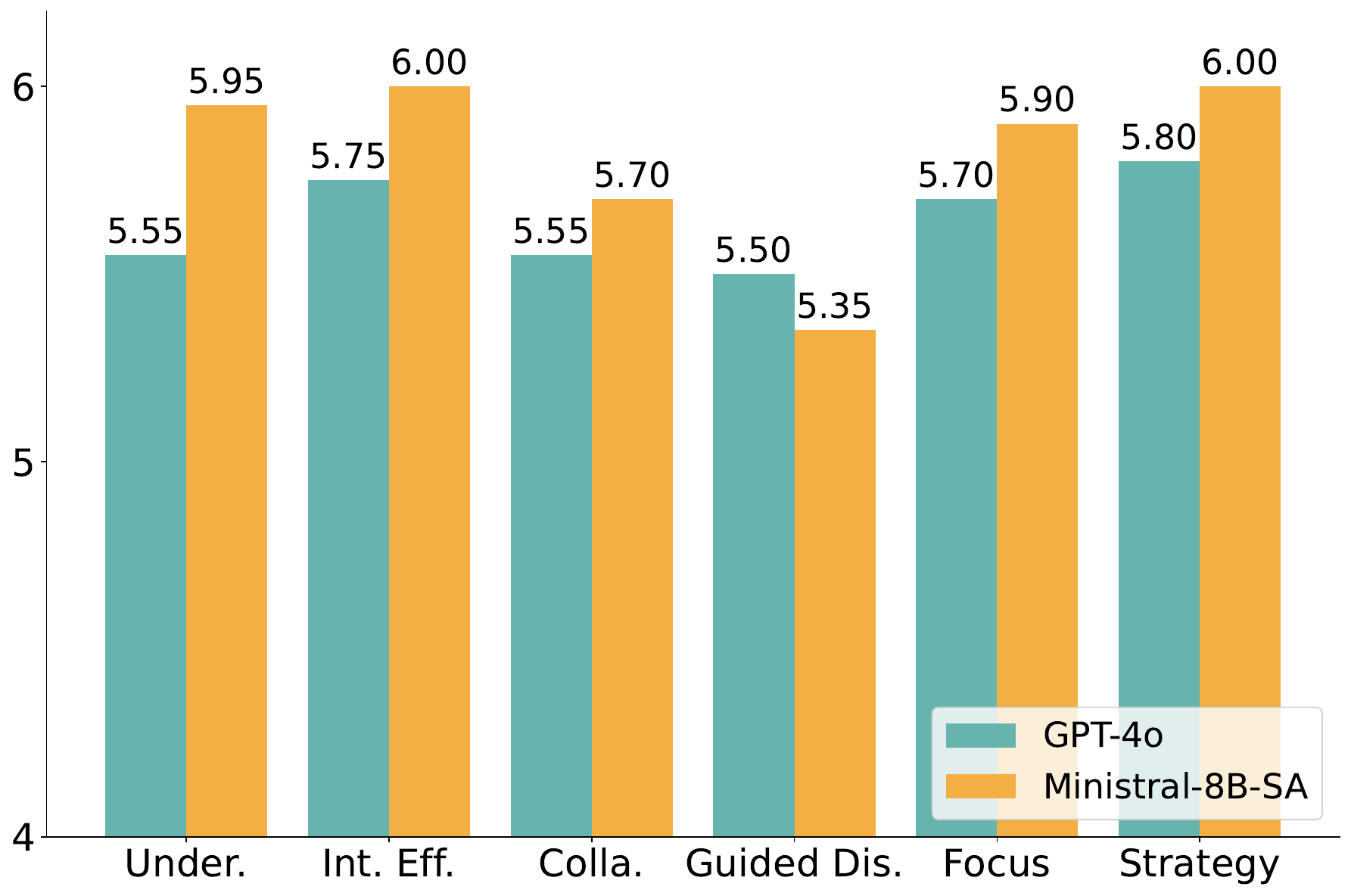}
    \caption{Results of Patient-$\Psi$-DM on general counseling and CBT-specific skills.}
    \label{fig:patient-psi}
    \vspace{-1mm}
\end{figure}

\begin{table}[t!]
    \centering
    \small
    \begin{tabular}{l|l|cc}
        \toprule
        \textbf{Setup} & \textbf{Model} & \textbf{Positive} ($\uparrow$) & \textbf{Negative} ($\downarrow$) \\
        \midrule
        Before & $-$ & $2.04$ & $3.49$ \\
        \midrule
        \multirow{2}{*}{After} & GPT-4o & $2.97$ & $2.10$ \\
         & Ministral-8B-SA & $\mathbf{3.10}$ & $\mathbf{1.97}$ \\
        \bottomrule
    \end{tabular}
    \caption{Results of Patient-$\Psi$-DM on PANAS for GPT-4o and Ministral-8B-SA.}
    \label{tab:patient-psi}
    \vspace{-4mm}
\end{table}

We evaluate counseling performance using two key metrics: CTRS and PANAS (Positive and Negative Affect Schedule). PANAS \cite{Watson1988DevelopmentAV} measures the changes in the client’s emotional state before and after counseling over ten positive and ten negative emotions. The prompts used for evaluation are provided in Figure \ref{fig:panas-prompt}.
As shown in Figure \ref{fig:patient-psi} and Table \ref{tab:patient-psi}, both GPT-4o and Ministral-8B-SA demonstrate effective CBT performance; however, Ministral-8B-SA outperforms GPT-4o in both general counseling and CBT-specific skills, leading to more substantial improvements in clients' positive emotions while effectively alleviating negative emotions, highlighting the potential of strategy-aligned CBT in real-world therapeutic settings.

\subsection{Case Studies}

\begin{figure*}[t!]
    \centering
    \begin{subfigure}[b]{0.49\linewidth}
        \includegraphics[width=\textwidth]{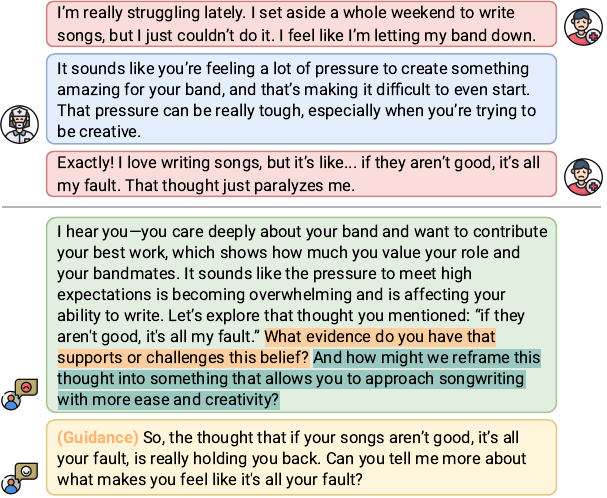}
        \caption{}
        \label{fig:ex1}
    \end{subfigure}
    \hfill
    \begin{subfigure}[b]{0.49\linewidth}
        \includegraphics[width=\textwidth]{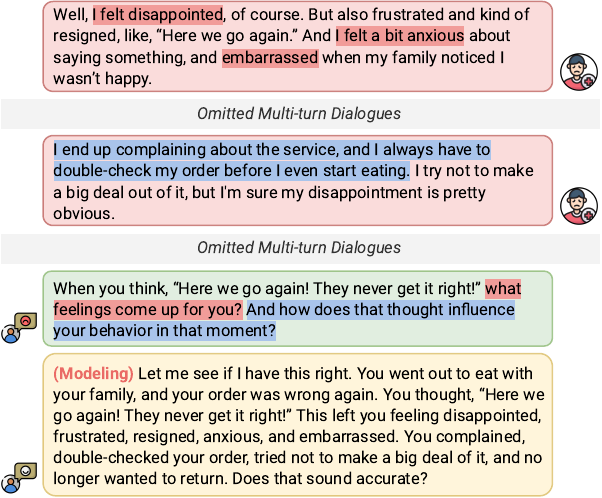}
        \caption{}
        \label{fig:ex2}
    \end{subfigure}
    \caption{Qualitative comparison of CBT counseling cases with and without strategy-aligned generation.}
    \label{fig:cases}
    \vspace{-2mm}
\end{figure*}

To provide an intuitive illustration, we qualitatively examine model responses conditioned on the same conversation histories. Our analysis indicates that models equipped with strategy-aligned generation adhere more consistently to the procedural structure of CBT. Specifically, they keep their questions focused on the immediate therapeutic objective, synthesize the information elicited into more complete formulations of clients’ negative feedback loops, and guide clients step by step toward independently developing alternative thoughts. In this section, we discuss two patterns that recur across the generated responses.

\paragraph{Strategy-Aligned Generation Improves Procedural Coherence.}

LLM-based counselors often fail to maintain procedural coherence when generation is not explicitly conditioned on counseling strategies. In particular, they may conflate multiple therapeutic objectives within a single turn, skip prerequisite stages before sufficient information has been elicited, and prematurely provide a ``correct'' interpretation before the client has fully articulated their experiences. As illustrated in Figure \ref{fig:cases}(a), the model without strategy alignment simultaneously asks \textit{what evidence supports or challenges this belief} and \textit{how the belief might be reframed}, collapsing exploratory guidance and cognitive restructuring, transitioning to intervention before the client's underlying thought has been sufficiently examined. In contrast, the strategy-aligned generation model selects ``\textit{Guidance}'' and restricts its response to eliciting the reasoning underlying the client’s feelings, thereby preserving the intended exploration function and giving the client a greater opportunity to articulate their own reasoning before alternative interpretations are introduced.

\paragraph{Strategy-Aligned Generation Prevents Redundant Elicitation.}

LLM-based counselors may also fail to recognize when sufficient information has already been elicited, leading them to revisit established events, thoughts, feelings, or behaviors, thereby stalling the progression of CBT. As exemplified in Figure \ref{fig:cases}(b), although the client has clearly articulated the negative thought, its associated feelings, and the resulting behaviors, the model redundantly asks about this information. In contrast, strategy-aligned generation appropriately transitions to the ``\textit{Modeling}'' stage, synthesizing the available information into a structured Events–Thoughts–Feelings–Behaviors formulation. By explicitly conditioning response generation on the appropriate counseling strategy, our approach mitigates redundant elicitation and facilitates a more coherent and procedurally grounded progression through the CBT session.

\section{Conclusion and Future Work}
\label{sec:conclusion}

We present \textsc{StratCBT}, the first dataset of mental health counseling with CBT strategies. It consists of $9,688$ sessions and $256$K utterances, with each counselor's response aligned with a distinct CBT strategy, surpassing existing datasets in general counseling and CBT-specific skills. Our experiments demonstrate the effectiveness of strategy-aligned generation and its potential in real-world settings.
In the future, we will encompass additional theories, such as SFBT, and improve performance through advanced planning techniques.

\section*{Limitations}

Although \textsc{StratCBT} advances the research by aligning dynamic psychological counseling with CBT strategy, its scope is currently limited to CBT and the English language because of the different strategy schemes in different treatment theories and the cross-cultural differences that may exist.
While CBT has been proven to be effective for ethnic minorities with diverse mental health problems \cite{annurev:/content/journals/10.1146/annurev-clinpsy-080921-072750}, its Western-centric approach leaves an open problem when generalizing to ethnic minorities due to some misunderstandings that may exist in terms of culture-specific information.
Therefore, we advocate for further research efforts to develop counseling conversation datasets with more diverse strategies and languages.

\section*{Ethical Considerations}

We discuss the following ethical considerations related to our \textsc{StratCBT} dataset:
(1) \textbf{Intellectual Property.} The original PatternReframe dataset is shared under the MIT licence\footnote{\url{https://mit-license.org/}}, which is free for research use. The construction of \textsc{StratCBT} is guided by counseling sessions from CheeseBurger Therapy. Although these sessions are publicly available, we will release only the \textsc{StratCBT} dataset and will not distribute any of the collected sessions. The entire construction process, including the modeled clients and synthetic dialogues, will be shared to guarantee interpretability.
(2) \textbf{Evaluator Treatments.} We hired postgraduate psychology students and fairly pay them according to the agreed instruction, salaries, and workloads.
(3) \textbf{Intended Use.} \textsc{StratCBT} can be utilized to develop more persuasive models in psychological counseling with CBT. Researchers can also inherit our dataset collection pipeline to develop their own datasets with additional treatment theories and languages.
(4) \textbf{Controlling Potential Risks.} Since the documents of \textsc{StratCBT} do not contain private information and the collection process is not necessary to make many judgments about social risks, we believe \textsc{StratCBT} does not introduce any additional risks. We manually verified a randomly sampled subset of the data to ensure the dataset did not contain any risky issues.
(5) \textbf{Use of AI.} We adopted ChatGPT for data analysis and sentence polishing. All outputs by LLMs were manually reviewed by authors.

\bibliography{custom}

\appendix

\onecolumn

\section{CBT Strategies}
\label{sec:strategies}

\begin{table*}[h]
    \centering
    \small
    \begin{tabular}{p{0.18\linewidth}|p{0.76\linewidth}}
        \toprule
        \textbf{Strategy} & \textbf{Definition \& Example} \\
        \midrule
        \multirow{3}{*}{\textit{Listening}} & Create a warm and empathetic environment through \textbf{restatements} and \textbf{validation}. \\
         & \textit{Example: It sounds like you’re having trouble with moving to a new city. That is a hard change for many people!} \\
        \midrule
        \multirow{3}{*}{\textit{Guidance}} & Encourage users to \textbf{explore} their \textbf{thoughts, feelings, and behaviors}. Draw connections between their values and their troubles. \\
         & \textit{Example: How did moving to a new city make you feel?} \\
        \midrule
        \multirow{6}{*}{\textit{Modeling}} & Assist users in categorizing their story as Event, Thoughts, Feelings, and Behavior, a fundamental chain reaction in CBT. \\
         & \textit{Example: Here's a summary I’ve put together based on what you've shared: -You moved to a new city, far from your friends and family -This led you to have the thought: ``I'm bad at interacting with new people'' -From these thoughts, you've felt significant anxiety and worry. -In response to these feelings, you've been avoiding leaving home.} \\
        \midrule
        \multirow{6}{*}{\textit{Story-telling}} & \textbf{Narrate} users' experience as Event $\rightarrow$ Thoughts $\rightarrow$ Feelings... Outline how this cycle forms a negative feedback loop. \\
         & \textit{Example: It's clear this has been a challenging event. You had envisioned a future in a new city after college, one v vibrant with friends and opportunities. However, you have had doubts about your ability to establish a new network of friends and are nervous about socializing. Because of this you are avoiding leaving your house. Does this way of explaining your experience ring true?} \\
        \midrule
        \multirow{7}{*}{\textit{Cognitive Reframing}} & Help users' \textbf{identify and apply cognitive distortions} that feel that their troubling thoughts most relate to. \\
         & \textit{Example: It sounds like you've been in a cycle fueled by conflicting desires. This is a common human predicament, and in CBT, we often look at these patterns through as \textbf{Cognitive Distortions}. In your situation, the distortion that might be most relevant is \textbf{All-or-Nothing thinking}. It's when we see things only in extremes – something is either perfect or a disaster, we're either a success or a failure. In your case [...]} \\
        \midrule
        \multirow{5}{*}{\textit{New Thoughts}} & Encourages users to create \textbf{new, positive thoughts} rooted in their main values identified. \\
         & \textit{Example: Excellent! Let's carry this insightful self-awareness forward. Now it's time to think about a new thought that can guide you in the face of similar decisions or situations. So putting yourself back in the situation where you are doubting your social skills, what's a new thought you might have to counter the ``All-or-Nothing'' thinking?} \\
        \midrule
        \multirow{4}{*}{\textit{Wrap Up}} & Help user reflect, and leave the session feeling changed, including (1) ask \textbf{user's feeling} compared to the beginning of the session and (2) remind user to \textbf{practice the thought} in future situations similar to the trouble they shared. \\
         & \textit{Example: How are you feeling now, compared to when we started?} \\
        \midrule
        \multirow{2}{*}{\textit{Others}} & Other utterances with actions not related to the aforementioned ones. \\
         & \textit{Example: Absolutely. I'm here for you whenever you're prepared.} \\
        \bottomrule
    \end{tabular}
    \caption{Definition of the CBT strategies in counseling conversations with examples.}
    \label{tab:strategies}
\end{table*}

\newpage

\twocolumn

\section{Evaluation Details}
\label{sec:evaluation}

\subsection{Human Correlation of CTRS Evaluation}
\label{sec:ctrs}

The correlations between human experts and GPT-4o on CTRS evaluation results, including Pearson ($r$), Spearman ($\rho$), and Kendall’s Tau ($\tau$) correlation coefficients, are listed in Table \ref{tab:correlation}.

\subsection{Baselines for Automatic Evaluation}
\label{sec:baselines}

\paragraph{Direct-Refine} is a straightforward refinement approach in which the model is directed to self-define and refine its initial response, guided by established counseling principles. Formally, the model $\mathcal{M}_{DR}$ directly generates a refined utterance $u_t^{ref}$ conditioned on the conversational history $\boldsymbol{u}$, the an initial system response $u_t^{sys}$, and the task instruction $I_{DR}$:
\begin{equation}
    u_t^{ref} = \mathcal{M}_{DR}(\boldsymbol{u}, u_t^{sys}, I_{DR}).
\end{equation}

\paragraph{Self-Refine} builds upon the approach initiated by \newcite{NEURIPS2023_91edff07}, where feedback focused on CBT principles is generated from the initial response, followed by a refinement of the response based on this feedback.
First, an LLM $\mathcal{M}_{SRF}$ generates feedback $f_t$ for the initial utterance. Subsequently, $\mathcal{M}_{SRR}$ produces a refined response based on this feedback and other contextual information:
\begin{align}
    f_t &= \mathcal{M}_{SRF}(\boldsymbol{u}, u_t^{sys}, I_{SRF}), \\
    u_t^{ref} &= \mathcal{M}_{SRR}(\boldsymbol{u}, u_t^{sys}, f_t, I_{SRR}).
\end{align}

\paragraph{Knowledge-Enhanced Generation} augments the generation process by incorporating commonsense causal knowledge relevant to the ongoing dialogue. For this, we select the CICERO v2 dataset \cite{shen2022multiviewcontextualcommonsenseinference} as the in-context example repository, which performs multi-view reasoning to provide commonsense explanations for events in binary dialogues.
Based on the conversation history $\boldsymbol{u} = \{u_1^{cli}, u_1^{sys}, \ldots, u_t^{usr}\}$, we select the top $K$ most similar example dialogues from the repository as in-context examples.

First, we vectorize the conversation history $\boldsymbol{u}$ and example dialogues $D = \{d_1, d_2, \ldots, d_r\}$ ($r$ is the number of example dialogues) to obtain the representation of the conversation history $\boldsymbol{u}_{emb}$ and example dialogues $D_{emb} = \{d_{emb_1}, d_{emb_2}, \ldots, d_{emb_r}\}$.
Then, we compute the semantic similarity between the conversation history and each of the example dialogues $d_{emb_i}$ ($i \in [1,r]$) via cosine similarity:
\begin{equation}
    Sim(\boldsymbol{u}, d_i) = \frac{\boldsymbol{u}_{emb} \cdot d_{emb_i}}{||\boldsymbol{u}_{emb}|| \cdot ||d_{emb_i}||},
\end{equation}
where $Sim(\cdot)$ denotes the semantic similarity score. Based on the similarity scores, we select $K = 4$ example dialogues as in-context examples $IC_{\boldsymbol{u}}$:
\begin{equation}
    IC_{\boldsymbol{u}} = TopK(Sim(\boldsymbol{u}, d_i)), \ \forall i \in [1,r].
\end{equation}
Finally, the in-context examples $IC_{\boldsymbol{u}}$ are fed into the LLM to perform commonsense reasoning and generate causal commonsense sentences, which are then integrated into the response generation process.

\begin{table}[t]
    \centering
    \small
    \begin{tabular}{r|ccc}
        \toprule
         & $r$ & $\rho$ & $\tau$ \\
        \midrule
        General Counseling & $0.60$ & $0.19$ & $0.16$ \\
        CBT-specific & $0.65$ & $0.65$ & $0.61$ \\
        \bottomrule
    \end{tabular}
    \caption{Pearson ($r$), Spearman ($\rho$), and Kendall’s Tau ($\tau$) correlation coefficients between human experts and GPT-4o on CTRS evaluation results \cite{lee-etal-2024-cactus}.}
    \label{tab:correlation}
    \vspace{-4mm}
\end{table}

\subsection{Preference Evaluation using the Bradley-Terry Model}
\label{sec:bt}

\begin{table*}[t!]
    \centering
    \small
    \begin{tabular}{l|c|cccccccc}
        \toprule
        \multirow{2}{*}{\textbf{Model}} & \multirow{2}{*}{\textbf{Overall}} & \multicolumn{8}{c}{\textbf{Strategy-Specific Accuracy}}  \\
        \cmidrule{3-10}
         & & Listen. & Guid. & Model. & Story. & Cog. Refram. & New. & Wrap Up & Others \\
        \midrule
        \rowcolor{gray!10}
        \multicolumn{10}{c}{\textit{Zero-shot LLMs}} \\
        \midrule
        \href{https://developers.openai.com/api/docs/models/gpt-4o}{GPT-4o} & $36.2$ & $28.6$ & $24.5$ & $48.5$ & $5.0$ & $37.1$ & $61.3$ & $47.1$ & $0.0$ \\
        \href{https://developers.openai.com/api/docs/models/gpt-4o-mini}{GPT-4o-mini} & $36.9$ & $\mathbf{58.6}$ & $6.6$ & $26.4$ & $4.2$ & $50.8$ & $54.7$ & $\underline{49.6}$ & $0.0$ \\
        \href{https://ai.google.dev/gemini-api/docs/models/gemini-2.0-flash}{Gemini-2.0-Flash} & $\mathbf{55.1}$ & $\underline{45.1}$ & $25.4$ & $50.7$ & $\mathbf{86.6}$ & $\mathbf{79.5}$ & $\mathbf{86.2}$ & $49.4$ & $0.0$ \\
        \href{https://docs.mistral.ai/models/mistral-small-2-0-24-09}{Mistral-Small-2.0} & $38.6$ & $26.8$ & $\mathbf{30.2}$ & $\underline{62.7}$ & $6.6$ & $36.6$ & $60.5$ & $\mathbf{50.1}$ & $0.0$ \\
        \href{https://ai.meta.com/blog/llama-4-multimodal-intelligence/}{Llama-4-Maverick} & $\underline{40.2}$ & $25.0$ & $\underline{28.8}$ & $\mathbf{77.8}$ & $\underline{9.4}$ & $39.9$ & $\underline{65.0}$ & $48.4$ & $0.0$ \\
        \href{https://api-docs.deepseek.com/news/news250325/}{DeepSeek-V3} & $33.5$ & $25.0$ & $28.4$ & $35.7$ & $1.0$ & $\underline{51.1}$ & $42.9$ & $45.5$ & $0.0$ \\
        \midrule
        \rowcolor{gray!10}
        \multicolumn{10}{c}{\textit{Fine-tuned LLMs}} \\
        \midrule
        \href{https://ai.meta.com/blog/meta-llama-3-1/}{Llama3.1-8B} & $\mathbf{81.7}$ & $\underline{74.2}$ & $\mathbf{72.3}$ & $\mathbf{81.2}$ & $96.6$ & $82.9$ & $85.8$ & $\mathbf{93.5}$ & $\mathbf{91.3}$ \\
        \href{https://docs.mistral.ai/models/ministral-8b-24-1}{Ministral-8B} & $81.1$ & $71.7$ & $71.4$ & $\underline{80.6}$ & $\mathbf{98.2}$ & $\underline{84.4}$ & $86.0$ & $92.1$ & $82.6$ \\
        \href{https://falconllm.tii.ae/falcon3/index.html}{Falcon3-7B} & $80.9$ & $\mathbf{77.5}$ & $70.6$ & $77.5$ & $96.8$ & $83.4$ & $\underline{86.2}$ & $91.6$ & $78.3$  \\
        \href{https://qwen.ai/blog?id=qwen3}{Qwen3-4B} & $\underline{81.3}$ & $72.5$ & $71.4$ & $78.6$ & $\mathbf{98.2}$ & $\mathbf{84.7}$ & $\mathbf{86.8}$ & $92.8$ & $\underline{87.0}$ \\
        \href{https://qwen.ai/blog?id=qwen3}{Qwen3-8B} & $80.8$ & $73.2$ & $\underline{71.7}$ & $75.9$ & $97.8$ & $81.1$ & $85.3$ & $\mathbf{93.5}$ & $82.6$ \\
        \bottomrule
    \end{tabular}
    \caption{Strategy prediction results on \textsc{StratCBT}. The best and second-best overall performance, accompanied by each strategy, are highlighted in \textbf{bold} and \underline{underlined}, respectively.}
    \label{tab:pred_strategy}
    \vspace{-4mm}
\end{table*}

In this study, we adopt the approach by \newcite{kang-etal-2024-large} and employ the Bradley-Terry model to assess the preferences of LLMs across diverse CBT strategies.
Consider the probability $P(s_i > s_j)$, quantifying the preference of strategy $s_i$ over the ground-truth $s_j$, can be formally defined as follows:
\begin{equation}
    P(s_i > s_j) = \frac{p_{s_i}}{p_{s_i} + p_{s_j}},
\end{equation}
where we assign a numerical score $n_{s_i}$ to each strategy $s_i$ and define $p_{s_i} = e^{n_{s_i}}$, which enables the expression of $P(s_i > s_j)$ in terms of these scores.
The likelihood of the preference $\mathbf{P}$ under the Bradley-Terry model is represented by the following equation:
\begin{equation}
    \mathbf{P} = \prod_{ij} [P(s_i > s_j)]^{w_{ij}} = \prod_{ij} \left ( \frac{p_{s_i}}{p_{s_i} + p_{s_j}} \right )^{w_{ij}},
\end{equation}
where $w_{ij}$ denotes the total number of times where strategy $s_i$ is preferred over strategy $s_j$. This leads to the following log-likelihood expression:
\begin{align}
    \log \mathbf{P} &= \sum_{ij} w_{ij} \log \frac{p_{s_i}}{p_{s_i} + p_{s_j}} \nonumber \\
    &= \sum_{ij} w_{ij} \log p_{s_i} - \sum_{ij} w_{ij} \log (p_{s_i} + p_{s_j}).
\end{align}

As noted by \newcite{Zermelo1929DieBD}, this expression has only a unique maximum. Consequently, we can differentiate it with respect to $p_i$ for any $i$ and set the derivative to zero:
\begin{equation}
    \frac{1}{p_{s_i}} \sum_j w_{ij} - \sum_j \frac{w_{ij} + w_{ji}}{p_i +p_j} = 0.
\end{equation}

Building on the efficient algorithm introduced by \newcite{JMLR:v24:22-1086}, this algorithm can be rearranged as follows:
\begin{gather}
    \frac{1}{p_{s_i}} \sum_j w_{ij} \frac{p_{s_j}}{p_{s_i} + p_{s_j}} - \sum_j \frac{w_{ji}}{p_{s_i} + p_{s_j}} = 0, \\
    p_i = \frac{\sum_j (w_{ij} \cdot p_{s_j}) / (p_{s_i} + p_{s_j})}{\sum_j w_{ji} / (p_{s_i} + p_{s_j})}.
\end{gather}
At this stage, we finally arrive at the iterative algorithm for the Bradley-Terry model, which calculates the preference $p_{s_i}$ for each strategy $s_i$.

\paragraph{Evaluation Setup.}

In our experiments, we follow \newcite{kang-etal-2024-large} in initializing all values ($p_i$) to $1$ and iteratively updating these estimates over $k = 20$ iterations. After each iteration, we normalize the values by dividing them by their geometric mean to maintain stability and ensure convergence, which is represented as follows:
\begin{equation}
    p_{s_i} \leftarrow \frac{p_{s_i}'}{\left ( \prod_{j=1} p_j' \right )^{1/N}},
\end{equation}
where $N = 8$ represents the total number of strategies. Upon each iteration, the converged values of $p$ indicate the preference $p_{s_i}$ for strategy $s_i$.

\subsection{Full Results for CoT-based Generation}
\label{sec:cot}

To validate the capability of models in implicit, real-time mental state reasoning with strategy-aligned generation, we conduct a comparative analysis using GPT-4o mini, comparing its performance in strategy-aligned generation utilizing internal reasoning against an explicit CoT that instructs the model to first generate a reasoning chain that includes the client's event, thought, feelings, and behaviors, followed by inferring the strategy before response generation. As depicted in Table \ref{tab:cot}, the results demonstrate comparable performance between explicit and implicit reasoning, suggesting that strategy-aligned generation effectively models clients' mental states.
Beyond supporting implicit mental state reasoning, strategy-aligned generation may also be viewed as a structured intervention on the response distribution. This perspective is reminiscent of perturbation-based learning, in that both use controlled variations to shape model behavior, an approach that has shown empirical effectiveness across diverse learning settings \cite{chen2026distributionally}.

\begin{table}[t!]
    \centering
    \small
    \begin{tabular}{l|ccc}
        \toprule
        \textbf{Model} & \textbf{BLEU} & \textbf{ROUGE-L} & \textbf{BERTScore} \\
        \midrule
        Standard & $5.7$ & $\underline{23.2}$ & $\underline{87.8}$ \\
        w/ Direct-Refine & $4.1$ & $21.1$ & $87.4$ \\
        w/ Self-Refine & $3.5$ & $19.9$ & $86.7$ \\
        w/ Knowledge & $5.4$ & $22.7$ & $\underline{87.8}$ \\
        Strategy-Aligned & $\underline{7.3}$ & $\mathbf{24.5}$ & $\mathbf{87.9}$ \\
        \rowcolor{gray!10}
        \quad \textit{w/ CoT} & $\mathbf{7.5}$ & $\mathbf{24.5}$ & $\underline{87.8}$ \\
        \bottomrule
    \end{tabular}
    \caption{Comparison of strategy-aligned generation between implicit reasoning and explicit CoT against baselines. The best and the second-best performance for each model and metric are in \textbf{bold} and \underline{underlined}.}
    \label{tab:cot}
    \vspace{-4mm}
\end{table}

\subsection{Full Results for Strategy Prediction}

The complete results for strategy prediction, including both zero-shot and fine-tuned LLMs, are presented in Table \ref{tab:pred_strategy}.

\onecolumn

\section{Prompts}

\begin{figure*}[h!]
    \begin{tcolorbox}[title=Prompt for Client Modeling, left=2mm,right=1mm,top=0mm, bottom=0mm,colback=white,colframe=CoolAccent]
    \begin{lstlisting}[style=plain]
You are a mental health counselor expert in Cognitive Behavioral Therapy (CBT). Your task is to reconstruct the event, feelings, and behaviors of a patient based on his/her negative thought.

Two examples are as follows:
{In-context Examples}

Please reconstruct the event, feelings, and behaviors of a patient based on his/her negative thought, ensure that the generated content is logically aligned with the thought and the given persona and consistent in actual experience.
Persona: {Percona}
Thought: {Thought}
    \end{lstlisting}
    \end{tcolorbox}

    \begin{tcolorbox}[title=Prompt for Dialogue Generation, left=2mm,right=1mm,top=0mm, bottom=0mm,colback=white,colframe=CoolAccent]
    \begin{lstlisting}[style=plain]
## Task Descriptions
You are a mental health counselor expert in Cognitive Behavioral Therapy (CBT). Your task is to reconstruct the CBT dialogue between a counselor and a client based on given event, thought, feelings, behaviors, and new thought.
Generally, the goal of a session include:
- Validate the client's trouble and values through listening and restatements.
- Understand the situation by identifying its surrounding events, thoughts, feelings, and behaviors.
- Identify problematic parts of thoughts and show the client why they are unhelpful.
- Create a new, more helpful thought that the client can repeat in similar circumstances.

## CBT Strategies
You should adhere and determine the correct CBT strategy when generating counselor's utterances. The strategy should be one of the following:
{Strategies}

## Process Flow of CBT
{Process Flow of CBT}

## Example Dialogue
An example is listed as follows:
Event: {Example Event}
Thought: {Example Thought}
Feelings: {Example Feelings}
Behaviors: {Example Behaviors}
New Thought: {Example New Thought}
Dialogue:
{Example Dialogue}

Please reconstruct a new CBT dialogue based on the following event, thought, feelings, behaviors, and new thought, and also determine the correct CBT strategy when generating the counselor's utterances. **Do not generate any strategy for the client's utterances.**
The dialogue should align with the client's persona, which is: {Persona}
The number of utterances should be no less than 20. You have to make the conversation gradually, making it as a long conversation. Make sure the dialogue starts with the client and ends with the counselor.
Event: {Event}
Thought: {Thought}
Feelings: {Feelings}
Behaviors: {Behaviors}
New Thought: {New Thought}
Dialogue:
    \end{lstlisting}
    \end{tcolorbox}
    \caption{Prompts for dataset construction, including client modeling and dialogue generation.}
    \label{fig:generation-prompt}
\end{figure*}

\begin{figure*}[h!]
    \begin{tcolorbox}[title=Prompt for Understanding Evaluation, left=2mm,right=1mm,top=0mm, bottom=0mm,colback=white,colframe=MutedGreen]
    \begin{lstlisting}[style=plain]
I want you to act as an evaluator. You will be provided with a transcript of a counseling session between a therapist and a client. [...]

[Evaluation Question]
How accurately does the therapist demonstrate understanding of the client's issues and concerns?

[criteria]
Score 0: Therapist repeatedly failed to understand what the patient explicitly said and thus consistently missed the point. Poor empathic skills.
Score 2: Therapist was usually able to reflect or rephrase what the patient explicitly said, but repeatedly failed to respond to more subtle communication. Limited ability to listen and empathize.
Score 4: Therapist generally seemed to grasp the patient's "internal reality" as reflected by both what the patient explicitly said and what the patient communicated in more subtle ways. Good ability to listen and empathize.
Score 6: Therapist seemed to understand the patient's "internal reality" thoroughly and was adept at communicating this understanding through appropriate verbal and non-verbal responses to the patient (e.g., the tone of the therapist's response conveyed a sympathetic understanding of the client's "message"). Excellent listening and empathic skills.
    \end{lstlisting}
    \end{tcolorbox}

    \begin{tcolorbox}[title=Prompt for Interpersonal Effectiveness Evaluation, left=2mm,right=1mm,top=0mm, bottom=0mm,colback=white,colframe=MutedGreen]
    \begin{lstlisting}[style=plain]
I want you to act as an evaluator. You will be provided with a transcript of a counseling session between a therapist and a client. [...]

[Evaluation Question]
How effective is the therapist in maintaining a positive and therapeutic relationship with the client?

[Criteria]
Score 0: Therapist had poor interpersonal skills. Seemed hostile, demeaning, or in some other way destructive to the patient.
Score 2: Therapist did not seem destructive, but had significant interpersonal problems. At times, therapist appeared unnecessarily impatient, aloof, insincere or had difficulty conveying confidence and competence.
Score 4: Therapist displayed a satisfactory degree of warmth, concern, confidence, genuineness, and professionalism. No significant interpersonal problems.
Score 6: Therapist displayed optimal levels of warmth, concern, confidence, genuineness, and professionalism, appropriate for this particular patient in this session.
    \end{lstlisting}
    \end{tcolorbox}

    \begin{tcolorbox}[title=Prompt for Collaboration Evaluation, left=2mm,right=1mm,top=0mm, bottom=0mm,colback=white,colframe=MutedGreen]
    \begin{lstlisting}[style=plain]
I want you to act as an evaluator. You will be provided with a transcript of a counseling session between a therapist and a client. [...]

[Evaluation Question]
To what extent does the therapist engage the client in collaborative goal-setting and decision-making?

[Criteria]
Score 0: Therapist did not attempt to set up a collaboration with patient.
Score 2: Therapist attempted to collaborate with patient, but had difficulty either defining a problem that the patient considered important or establishing rapport.
Score 4: Therapist was able to collaborate with patient, focus on a problem that both patient and therapist considered important, and establish rapport.
Score 6: Collaboration seemed excellent; therapist encouraged patient as much as possible to take an active role during the session (e.g., by offering choices) so they could function as a "team".
    \end{lstlisting}
    \end{tcolorbox}
    \caption{Prompts for evaluating CTRS for general counseling skills \cite{lee-etal-2024-cactus}.}
    \label{fig:ctrs-prompt}
\end{figure*}

\begin{figure*}[h!]
    \begin{tcolorbox}[title=Prompt for Guided Discovery Evaluation, left=2mm,right=1mm,top=0mm, bottom=0mm,colback=white,colframe=MutedGreen]
    \begin{lstlisting}[style=plain]
I want you to act as an evaluator. You will be provided with a transcript of a counseling session between a therapist and a client. [...]

[Evaluation Question]
How effectively does the therapist use guided discovery techniques to facilitate client self-reflection and insight?

[criteria]
Score 0: Therapist relied primarily on debate, persuasion, or "lecturing." Therapist seemed to be "cross-examining" patient, putting the patient on the defensive, or forcing his/her point of view on the patient.
Score 2: Therapist relied too heavily on persuasion and debate, rather than guided discovery. However, therapist's style was supportive enough that patient did not seem to feel attacked or defensive.
Score 4: Therapist, for the most part, helped patient see new perspectives through guided discovery (e.g., examining evidence, considering alternatives, weighing advantages and disadvantages) rather than through debate. Used questioning appropriately.
Score 6: Therapist was especially adept at using guided discovery during the session to explore problems and help patient draw his/her own conclusions. Achieved an excellent balance between skillful questioning and other modes of intervention.
    \end{lstlisting}
    \end{tcolorbox}

    \begin{tcolorbox}[title=Prompt for Focus Evaluation, left=2mm,right=1mm,top=0mm, bottom=0mm,colback=white,colframe=MutedGreen]
    \begin{lstlisting}[style=plain]
I want you to act as an evaluator. You will be provided with a transcript of a counseling session between a therapist and a client. [...]

[Evaluation Question]
How well does the therapist identify and address the client's key cognitions or behaviors that need change?

[criteria]
Score 0: Therapist did not attempt to elicit specific thoughts, assumptions, images, meanings, or behaviors.
Score 2: Therapist used appropriate techniques to elicit cognitions or behaviors; however, therapist had difficulty finding a focus or focused on cognitions/behaviors that were irrelevant to the patient's key problems.
Score 4: Therapist focused on specific cognitions or behaviors relevant to the target problem. However, therapist could have focused on more central cognitions or behaviors that offered greater promise for progress.
Score 6: Therapist very skillfully focused on key thoughts, assumptions, behaviors, etc. that were most relevant to the problem area and offered considerable promise for progress. 
    \end{lstlisting}
    \end{tcolorbox}

    \begin{tcolorbox}[title=Prompt for Strategy Evaluation, left=2mm,right=1mm,top=0mm, bottom=0mm,colback=white,colframe=MutedGreen]
    \begin{lstlisting}[style=plain]
I want you to act as an evaluator. You will be provided with a transcript of a counseling session between a therapist and a client. [...]

[Evaluation Question]
How appropriate and coherent is the therapist's strategy for promoting change in the client's problematic behaviors or thoughts?

[criteria]
Score 0: Therapist did not select cognitive-behavioral techniques.
Score 2: Therapist selected cognitive-behavioral techniques; however, either the overall strategy for bringing about change seemed vague or did not seem promising in helping the patient.
Score 4: Therapist seemed to have a generally coherent strategy for change that showed reasonable promise and incorporated cognitive-behavioral techniques.
Score 6: Therapist followed a consistent strategy for change that seemed very promising and incorporated the most appropriate cognitive-behavioral techniques.
    \end{lstlisting}
    \end{tcolorbox}
    \caption{Prompts for evaluating CTRS for CBT-specific skills \cite{lee-etal-2024-cactus}.}
    \label{fig:ctrs-prompt-2}
\end{figure*}

\begin{figure*}[h!]
    \begin{tcolorbox}[title=Prompt for PANAS Evaluation Before Counseling, left=2mm,right=1mm,top=0mm, bottom=0mm,colback=white,colframe=WarmOrange]
    \begin{lstlisting}[style=plain]
A person with the characteristics listed in the intake form received counseling. Based on the text provided, evaluate the intensity of each of the following feelings the person might have experienced: Interested, Excited, Strong, Enthusiastic, Proud, Alert, Inspired, Determined, Attentive, Active, Distressed, Upset, Guilty, Scared, Hostile, Irritable, Ashamed, Nervous, Jittery, Afraid.

For each feeling, generate a score from 1 to 5 using the following scale:
1 - Very slightly or not at all
2 - A little
3 - Moderately
4 - Quite a bit
5 - Extremely

Additionally, provide a brief explanation for each score. Separate feeling, explanation, score by comma. Do not add any prefix.

Here is the text:
{Cognitive Model}
    \end{lstlisting}
    \end{tcolorbox}

    \begin{tcolorbox}[title=Prompt for PANAS Evaluation After Counseling, left=2mm,right=1mm,top=0mm, bottom=0mm,colback=white,colframe=WarmOrange]
    \begin{lstlisting}[style=plain]
A person with the characteristics listed in the intake form received counseling. The following counseling session is a conversation between the client and the counselor. After reviewing the conversation, evaluate the intensity of each of the following feelings the person might have experienced once the counseling session is complete: Interested, Excited, Strong, Enthusiastic, Proud, Alert, Inspired, Determined, Attentive, Active, Distressed, Upset, Guilty, Scared, Hostile, Irritable, Ashamed, Nervous, Jittery, Afraid.

For each feeling, generate a score from 1 to 5 using the following scale:
1 - Very slightly or not at all
2 - A little
3 - Moderately
4 - Quite a bit
5 - Extremely

Additionally, provide a brief explanation for each score. Separate feeling, explanation, score by comma. Do not add any prefix.

Here is the text:
{Cognitive Model}

Here is the counseling session:
{Dialogue}
    \end{lstlisting}
    \end{tcolorbox}
    \caption{Prompts for evaluating PANAS before and after counseling \cite{lee-etal-2024-cactus}.}
    \label{fig:panas-prompt}
\end{figure*}

\begin{figure*}[h!]
    \begin{tcolorbox}[title=Prompt for Standard Generation, left=2mm,right=1mm,top=0mm, bottom=0mm,colback=white,colframe=IAP]
    \begin{lstlisting}[style=plain]
You are a mental health counselor expert in Cognitive Behavioral Therapy (CBT). Your task is to generate the next counselor utterance in the counseling dialogue.
Generally, the goal of a session include:
- Validate the client's trouble and values through listening and restatements.
- Understand the situation by identifying its surrounding events, thoughts, feelings, and behaviors.
- Identify problematic parts of thoughts and show the client why they are unhelpful.
- Create a new, more helpful thought that the client can repeat in similar circumstances.

Counseling Dialogue:
{Dialogue}
    \end{lstlisting}
    \end{tcolorbox}

    \begin{tcolorbox}[title=Prompt for Strategy Prediction, left=2mm,right=1mm,top=0mm, bottom=0mm,colback=white,colframe=IAP]
    \begin{lstlisting}[style=plain]
You are a mental health counselor expert in Cognitive Behavioral Therapy (CBT). Your task is to determine the correct CBT strategy when generating the next counselor utterance in the counseling dialogue.
The strategy should be one of the following, output the name of the strategy only:
{Strategy Definitions}

Counseling Dialogue:
{Dialogue}
    \end{lstlisting}
    \end{tcolorbox}
    
    \begin{tcolorbox}[title=Prompt for Strategy-Aligned Generation, left=2mm,right=1mm,top=0mm, bottom=0mm,colback=white,colframe=IAP]
    \begin{lstlisting}[style=plain]
You are a mental health counselor expert in Cognitive Behavioral Therapy (CBT). Your task is to generate the next counselor utterance in the counseling dialogue.
Generally, the goal of a session include:
- Validate the client's trouble and values through listening and restatements.
- Understand the situation by identifying its surrounding events, thoughts, feelings, and behaviors.
- Identify problematic parts of thoughts and show the client why they are unhelpful.
- Create a new, more helpful thought that the client can repeat in similar circumstances.
You should follow the given strategy with generating the utterance.

Strategy:
{Selected Strategy Definition}

Counseling Dialogue:
{Dialogue}
    \end{lstlisting}
    \end{tcolorbox}
    \caption{Prompts for instructing zero-shot and fine-tuned models.}
    \label{fig:conversation-prompt}
\end{figure*}

\newpage

\end{document}